\documentclass[letterpaper, 10 pt, conference]{ieeeconf}  

\IEEEoverridecommandlockouts                              
\usepackage[utf8]{inputenc}
\usepackage[T1]{fontenc}
\usepackage{graphicx} 
\usepackage{amsmath} 
\usepackage{amssymb}  
\usepackage[style=ieee,backend=biber]{biblatex}
\usepackage{xcolor}
\bibliography{tiago_tactile}

\title{\LARGE \bf
Leveraging Touch Sensors to Improve Mobile Manipulation
}

\author{Luca Lach$^{1,2}$, Robert Haschke$^{2}$, Francesco Ferro$^{1}$, Jordi Pagès$^{1}$
\thanks{$^{1}$PAL Robotics S.L., Barcelona, Spain}%
\thanks{$^{2}$Neuroinformatics Group, Bielefeld University, Germany}
\thanks{}
\thanks{This work was supported by the European Union Horizon 2020 Marie}
\thanks{Curie Actions under Grant 813713 NeuTouch.}}%

\begin{document}

\maketitle
\thispagestyle{empty}
\pagestyle{empty}

\begin{abstract}

Despite many advances in service robotics, successful and secure object manipulation on mobile platforms is still a challenge. 
In order to come closer to human grasping performance, it is natural to provide robots with the same capability that humans have: the sense of touch.
This abstract presents novel, tactile-equipped end-effectors for the service robot TIAGo that are currently being developed. 
Their primary goal is to improve reliability and success of mobile manipulation, but they also enable further research in related fields such as learning by human demonstration, object exploration and force control algorithms.

\end{abstract}

\section{INTRODUCTION}
Moving forward to an era where robots are more integrated in society and able to assist humans more and more, identifying, studying and improving the different capabilities of robotic platforms is required.
This is crucial to have in order to perform tasks autonomously and effectively. 
When analyzing these capabilities, one of the challenges that mobile service robots face is the ability to securely grasp objects.
Reliable manipulation is a fundamental skill needed to improve interaction capabilities of service robots both with humans and their environment.
One of the major reasons why manipulation performance of mobile robots is currently unreliable is the lack of  feedback from the environment. 
With no information about the physical interaction of the end-effector with the environment, robots are not able to correct a poorly performed grasp, detect object slippage or even assess whether an object was grasped at all.

After vision, the sense of touch is the most important sensory modality that humans use to manipulate objects.
We use it  to detect initial object contact and acquire the information needed for hand/object interactions.
Additionally, we use it for perceiving variations in the contact itself like slippage, vibrations, or grip strength. 
Touch is also vital for robots which physically interact with objects and humans, to sense the properties of objects, learn how to use them, and enable cooperation.
This work presents two new end-effector configurations for TIAGo's two-finger gripper that provide tactile data to the robot.
Both of the presented configurations are designed for different kinds of tasks and applications.

\section{TIAGo ROBOT PLATFORM}

\begin{figure}[thpb]
  \centering
  \includegraphics[scale=0.35]{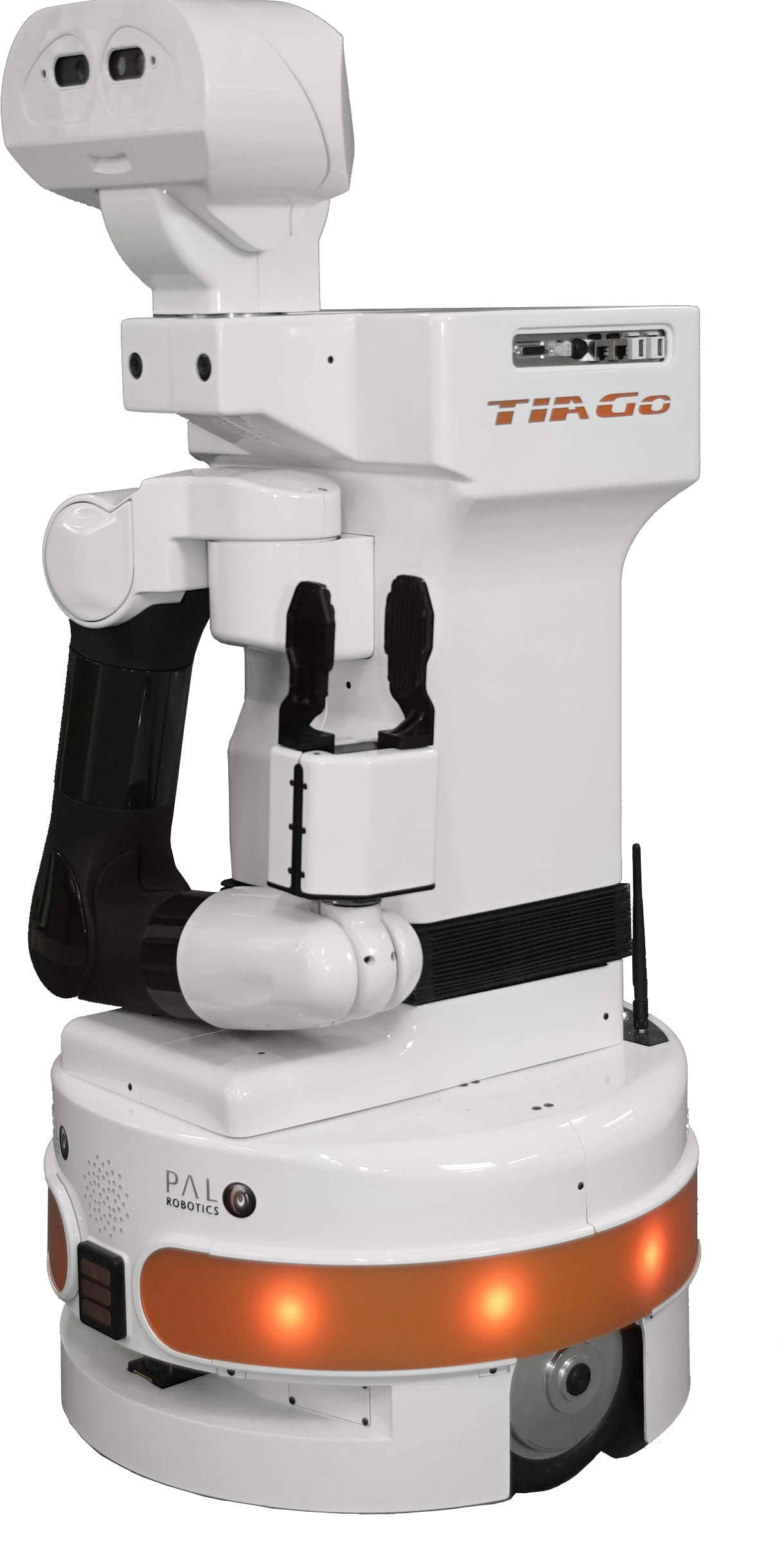}
  \caption{The TIAGo service robot with its 7-DoF arm and a two-finger parallel jaw gripper manipulator. Both fingers will be equipped with tactile sensors for the newly proposed gripper configurations.}
  \label{tiago}
\end{figure}

The TIAGo robot \cite{pages2016tiago} has a modular design, inspired by the technology of its predecessors REEM-C and REEM\cite{marchionni2013reem} and can be configured based on specific needs. 
TIAGo makes use of existing components and proposes a new modular architecture allowing the robot to adapt to different applications and expand its capabilities.
The arm has a large manipulation workspace, being able to reach the ground as well as high shelves. 
There are two different end-effectors for TIAGo, a two-finger gripper and a humanoid hand which can be quickly exchanged for performing various manipulation tasks.

For this work, the focus lies on the two-finger gripper that is illustrated in Figure \ref{tiago}.
Most basic manipulation tasks such as grasping and delivering objects can be accomplished with such a two-finger gripper.
Thanks to its modular design, it is straight forward to develop new kinds of fingers that can be mounted on the existing gripper base containing the motors and control electronics.
The first gripper configuration is simple, yet effective.
It replaces the non-sensing plastic fingers with strain gauge sensors, which provide one force measurement per finger.
These sensors are highly sensitive to touch, enabling quick first touch detection.
Their simplicity is another advantage in many use cases: by providing only one value regarding if and how much force is measured, integrating this information into existing control algorithms is simple and does not require pre-processing.

In a second version of the tactile gripper, the fingers will be equipped with a touch sensor matrix, similar to the one presented in \cite{schopfer2012slippage}.
These types of sensors provide multiple touch readings per finger at high frequencies.
While they deliver larger amounts of data, thus requiring more  pre-processing, they also enable a more complex analysis of the physical interaction.
Furthermore, the sensor output can be represented as a grey scale image, making it well-suited for machine learning techniques such as Convolutional Neural Networks (CNNs) \cite{krizhevsky2012imagenet}.

\begin{figure}[thpb]
  \centering
  \includegraphics[scale=0.3]{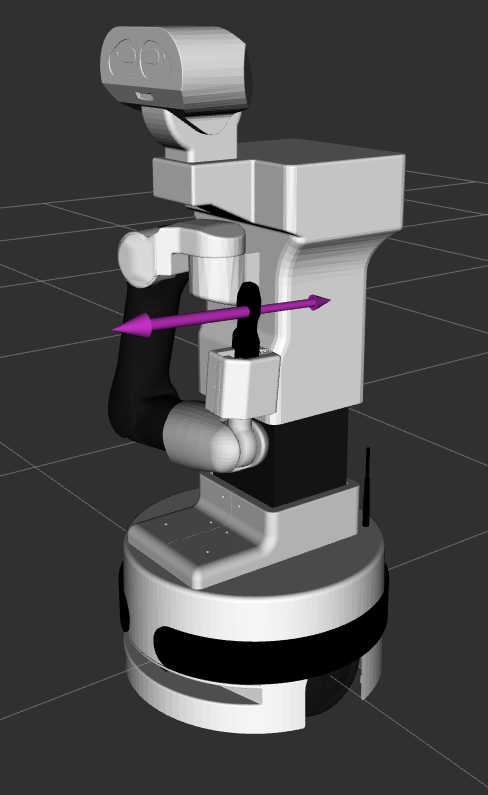}
  \caption{TIAGo in rviz with a visualization of forces measured at its fingers using strain gauge sensors. The arrow sizes indicate the force magnitudes.}
  \label{tiago_force}
\end{figure}

\section{APPLICATIONS OF TOUCH SENSORS IN MANIPULATION}

Current state-of-the-art object grasping algorithms essentially perform blind manipulation, which is executing the grasp without any kind of feedback from the environment.
As a result, general manipulation performance of mobile robots is rather poor.
The two previously introduced gripper configurations provide a solution to this problem by employing touch sensors on the fingers.

The configuration using two strain gauge sensors is designed to be easily integrated with current manipulation algorithms.
Figure \ref{tiago_force} shows a visualization of the measured forces in rviz, a visualization tool of ROS \cite{quigley2009ros}.
By simply determining a noise threshold and a maximum desired force, one can improve existing control algorithms with using force control.
The main advantage of using force instead of position or velocity control is that the robot can reliably measure whether it successfully acquired object contact.
Upon contact, it is then able to decide with how much force to handle the object to manipulate.
In situations where the robot needs to handle very fragile objects, it might choose lower forces to take special care not to damage the object.
This simple gripper modification can already increase manipulation performance in most basic mobile grasping tasks.

For more advanced manipulation tasks, the second gripper version will be equipped with sensor matrices.
Acquiring more detailed information about the physical interaction with the object enables the robot to accomplish more complex tasks.
One possible issue with object manipulation which often arises after the actual grasp is slippage.
With the object in the gripper, movements of the arm or of the robot itself can cause object slippage which can eventually lead to the object slipping out of the end-effector.
When using touch sensor matrices, it was shown that slippage can be detected early enough using CNNs \cite{meier2016tactile} to readjust the grip. 
Another scenario where touch sensors can improve robot performance is learning by human demonstration. 
Typically, robots are taught example trajectories which they can either repeat or learn to adapt them to new situations \cite{argall2009survey}.
When these trajectories involve physical interaction with objects, the robot has no way of knowing this.
However, if during the recording of the samples the forces are recorded and the robot has tactile sensors, it can replicate how the object is handled.
This would ensure safe handling of objects and aim to minimize damages.

\section{CONCLUSION}

This paper has presented the challenges that the robots face when grasping and manipulating objects and how tactile sensors improve upon this issue. 
In order to improve classical control algorithms, a gripper with two strain gauges was proposed.
For more complex tasks which may also involve machine learning, a second gripper configuration with sensor matrices on the fingers was proposed.
The grippers are developed for and will be tested on the service robot TIAGo, a capable mobile platform.
Using TIAGo, the hardware can easily and quickly be tested in a variety of real-world scenarios.

\vspace{0.5cm}
\renewcommand*{\UrlFont}{\rmfamily}
\printbibliography

\addtolength{\textheight}{-12cm}   

\end{document}